\documentclass[10pt,twocolumn,letterpaper]{article}

\usepackage{cvpr}
\usepackage{times}
\usepackage{epsfig}
\usepackage{graphicx}
\usepackage{amsmath}
\usepackage{amssymb}
\usepackage{listings}
\usepackage{amsmath}
\usepackage{amssymb}
\usepackage{pbox}
\usepackage{makecell}

\usepackage{algorithm}
\usepackage{algpseudocode}

\usepackage{epsfig}
\usepackage{graphicx}
\usepackage{subcaption}

\usepackage{booktabs}
\usepackage{pifont}
\newcommand{\cmark}{\ding{51}}%
\newcommand{\xmark}{\ding{55}}%

\usepackage[table,x11names]{xcolor}

\usepackage{multicol}
\usepackage{multirow}
\usepackage[misc]{ifsym}

\newcommand{\comment}[1]{}

\definecolor{LightCyan}{rgb}{0.88,1,1}

\newlength\savewidth\newcommand\shline{\noalign{\global\savewidth\arrayrulewidth
  \global\arrayrulewidth 1pt}\hline\noalign{\global\arrayrulewidth\savewidth}}

\makeatletter
\def\@fnsymbol#1{\ensuremath{\ifcase#1\or \dagger\or \dagger\or
\mathsection\or \mathparagraph\or \|\or **\or \dagger\dagger
\or \ddagger\ddagger \else\@ctrerr\fi}}
\makeatother

\usepackage[pagebackref=true,breaklinks=true,letterpaper=true,colorlinks,bookmarks=false]{hyperref}

\usepackage{algorithm}
\usepackage[]{algpseudocode}

\cvprfinalcopy 

\begin{document}

\title{Mask Frozen-DETR: High Quality Instance Segmentation with One GPU}


\author{
  Zhanhao Liang \quad\quad\quad\quad\quad\quad\quad\quad\quad Yuhui Yuan\thanks{Corresponding author. \Letter\space \texttt{yuhui.yuan@microsoft.com}} \\
  The Australian National University\quad\quad\quad
  Microsoft Research Asia \\
}

\maketitle

\begin{abstract}
In this paper, we aim to study how to build a strong instance segmenter with minimal training time and GPUs, as opposed to the majority of current approaches that pursue more accurate instance segmenter by building more advanced frameworks at the cost of longer training time and higher GPU requirements. To achieve this, we introduce a simple and general framework, termed Mask Frozen-DETR, which can convert any existing DETR-based object detection model into a powerful instance segmentation model. Our method only requires training an additional lightweight mask network that predicts instance masks within the bounding boxes given by a frozen DETR-based object detector. Remarkably, our method outperforms the state-of-the-art instance segmentation method Mask DINO in terms of performance on the COCO test-dev split (55.3\% vs. 54.7\%) while being over 10$\times$ times faster to train. Furthermore, all of our experiments can be trained using only one Tesla V100 GPU with 16 GB of memory, demonstrating the significant efficiency of our proposed framework.
\end{abstract}

\section{Introduction}

Instance segmentation is one of the most fundamental but difficult computer vision tasks requiring pixel-level localization and recognition in the given input image. Most advances in modern image instance segmentation methods are heavily influenced by the latest state-of-the-art 2D object detection systems. For example, two representative leading instance segmentation approaches, including Cascade Mask R-CNN~\cite{cai2019cascade} and Mask DINO~\cite{li2022mask}, are built by adding a parallel segmentation branch to the strong object detection systems, specifically Cascade R-CNN~\cite{cai2018cascade} and DINO~\cite{zhang2022dino}.

\begin{figure}[t]
\centering
\begin{minipage}[t]{1\linewidth}
\centering
\begin{subfigure}[b]{\linewidth}
\centering
\includegraphics[height=0.3\columnwidth]{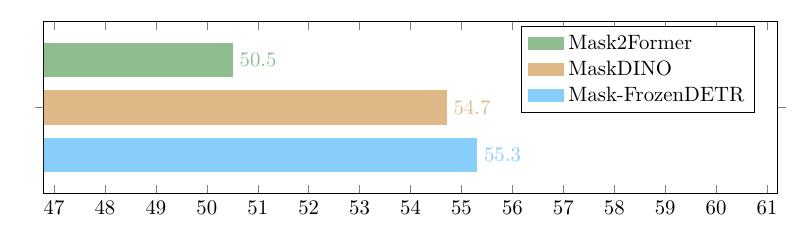}
\caption{mask AP}
\end{subfigure}
\hfill
\begin{subfigure}[b]{\linewidth}
\centering
\includegraphics[height=0.3\columnwidth]{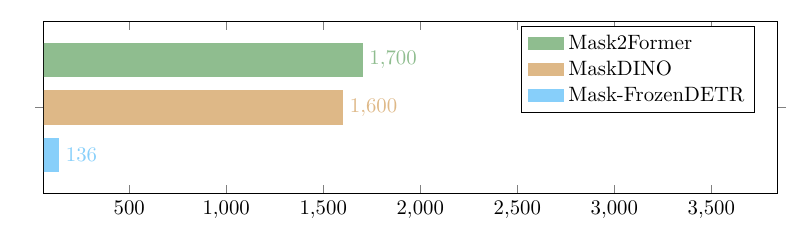}
\caption{GPU Hours}
\end{subfigure}
\end{minipage}
\caption{\small{Illustrating the comparison results with Mask2Former and Mask DINO on COCO instance segmentation task. (a) our Mask Frozen-DETR outperforms the previous SOTA Mask DINO by +$0.5\%$. (b) our Mask Frozen-DETR speeds up the training by more than $10\times$ times compared to Mask DINO.
}}
\label{fig:attention_maps_intro}
\end{figure}

\begin{figure*}[t]
\begin{minipage}[t]{1\linewidth}
\centering
\begin{subfigure}[b]{1\textwidth}
\centering
\includegraphics[width=1\textwidth]{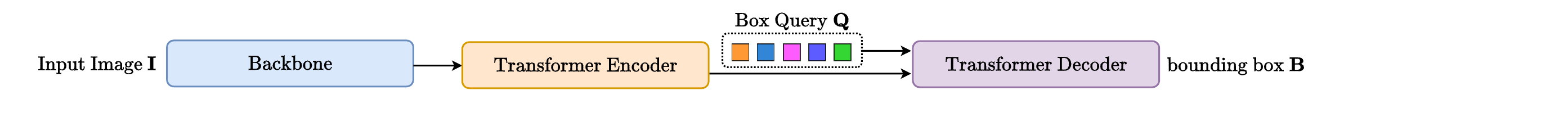}
\vspace{-3mm}
\caption{\footnotesize{DETR-based object detection framework}}
\end{subfigure}
\begin{subfigure}[b]{1\textwidth}
\centering
\vspace{3mm}
\includegraphics[width=1\textwidth]{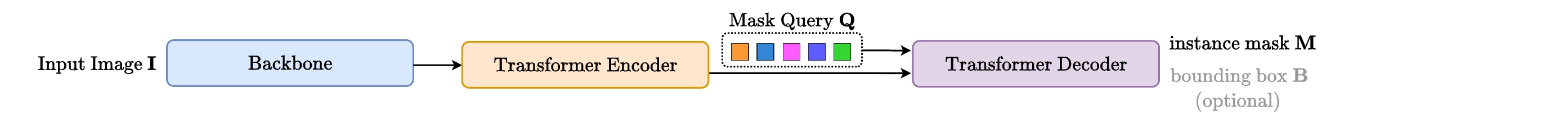}
\vspace{-5mm}
\caption{\footnotesize{DETR-based instance segmentation framework}}
\end{subfigure}
\begin{subfigure}[b]{1\textwidth}
\centering
\vspace{3mm}
\includegraphics[width=1\textwidth]{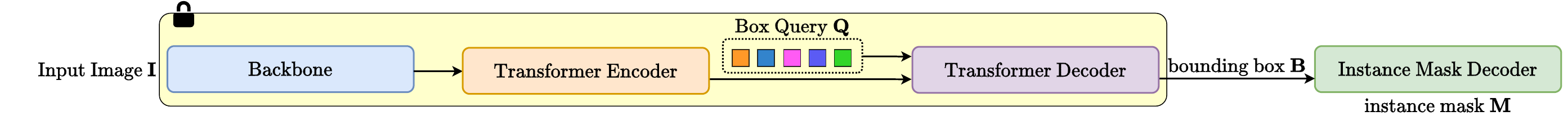}
\vspace{-3mm}
\caption{\footnotesize{Our approach: Mask Frozen-DETR}}
\end{subfigure}
\end{minipage}
\caption{\small{Illustrating the overall pipelines of DETR-based object detectors ($1$-st row), instance segmenters ($2$-ed row) based on DETR, and the proposed approach. Instead of training the instance segmentation models from scratch, we propose a Mask Frozen-DETR that uses a frozen DETR-based object detector to generate the bounding boxes and then trains a mask head to produce instance segmentation masks.}}
\label{fig:intro}
\end{figure*}

Despite the convergence of deep neural network architectures between object detection and instance segmentation, most existing efforts still need independent training based on supervision signals of different granularity, i.e., bounding boxes vs. instance masks. Training modern instance segmentation models from scratch is resource-intensive and time-consuming. For example, Using ResNet-$50$~\cite{He2016ResNet} and Swin-L~\cite{liu2021swin} as backbone networks, Mask2Former~\cite{cheng2022masked} requires over $500\times$ and $1700\times$ V100 GPU hours for training, respectively.

We show that the existing DETR-based object detection models can be efficiently converted into strong instance segmentation models, unlike the previous efforts that train the instance segmentation models from scratch. We start from the recent powerful 2D object detection models, i.e., $\mathcal{H}$-DETR~\cite{jia2022detrs} and DINO-DETR~\cite{zhang2022dino}. We propose two key innovations: (i) design a light-weight instance segmentation network that effectively uses the output of a frozen DETR-based object detector, including the object query and the encoder feature map, to predict instance masks, and (ii) demonstrate the high training efficiency of our approach compared to the previous state-of-the-art instance segmentation approaches while achieving competitive performance under different model scales.

We conduct comprehensive comparison experiments on the COCO instance segmentation benchmark to verify the effectiveness of our approach. Our approach achieves strong results with a very short learning time. For instance, with only $6\times$ training epochs, our approach slightly surpasses the state-of-the-art method Mask DINO. Remarkably, the entire training process of our approach takes less than $140\times$ GPU hours, while Mask DINO takes nearly $1600\times$ GPU hours. Consequently, our approach improves the training efficiency by more than $10\times$. We hope our simple approach can enable broader research communities to contribute to advancing stronger instance segmentation models.

\section{Related Work}

\vspace{1mm}
\noindent\textbf{Object Detection.}
Object detection is a fundamental research area that has produced a lot of excellent work, such as Faster R-CNN~\cite{ren2015faster}, Cascade R-CNN~\cite{cai2018cascade}, YOLO~\cite{redmon2016you}, DETR~\cite{carion2020end}, and Deformable DETR~\cite{zhu2020deformable}.
Recently, most of the exciting progress in object detection mainly comes from the developments of various DETR-based approaches, including DINO-DETR~\cite{zhang2022dino} and $\mathcal{H}$-DETR~\cite{jia2022detrs}.
The current state-of-the-art methods~\cite{wang2022internimage,lin2023detr,ma2023revisiting,liang2022expediting,he2022rankseg} are also built based on them.
In general, we can easily access a lot of object detection model weights based on DETR framework as most of them are open-sourced.
We show our approach can be extended to these modern DETR-based detectors easily and achieves strong performance while being more than $10\times$ faster to train by exploiting the off-the-shelf pre-trained weights on object detection tasks.

\vspace{1mm}
\noindent\textbf{Instance Segmentation.}
Instance segmentation is a computer vision task that requires an algorithm to assign a pixel-level or point-level mask with a category label for each instance of interest in an image, video or point cloud. Most existing methods follow the R-CNN~\cite{girshick2014rich} paradigm, which first detects objects and then segments them. For example, Mask R-CNN~\cite{he2017mask} extends Faster R-CNN~\cite{ren2015faster} with a fully convolutional mask head, Casacde Mask R-CNN combines Casacde R-CNN~\cite{cai2018cascade} with Mask R-CNN, and HTC~\cite{chen2019hybrid} improves the performance with interleaved execution and mask information flow. Some recent methods propose more concise designs such as SOLO~\cite{xwang20} that segments objects by locations without bounding boxes or embedding learning, QueryInst~\cite{FangQueryInst} that performs end-to-end instance segmentation based on Sparse RCNN~\cite{sun2021sparse}, MaskFormer~\cite{cheng2021per} and Mask2Former\cite{cheng2022masked} that use a simple mask classification based on DETR~\cite{carion2020end}, and Mask-DINO~\cite{li2022mask} that extends DINO by adding a mask prediction branch that supports all image segmentation tasks using query embeddings and pixel embeddings.
Moreover, the very recent Mask3D~\cite{schult2022mask3d} and SPFormer~\cite{sun2022superpoint} have built state-of-the-art 3D instance segmentation systems following the design of Mask2Former.

\noindent\textbf{Discussion.}
Most of the existing efforts train the instance segmentation models from scratch without using the off-the-shelf object detection model weights, thus requiring very expensive training.
Our approach uses the weights of frozen DETR-based models and introduces a very efficient instance segmentation head design with high training efficiency.
Figure~\ref{fig:intro} illustrates the differences between the existing methods and our approach.
For example, the very recent state-of-the-art Mask DINO is trained from scratch while we simply freeze the existing object detector and train a very light instance mask decoder.
We empirically show the great advantages of our approach on the COCO instance segmentation task with experiments under various settings.
Notably, our approach sets new records on the challenging COCO instance segmentation task while accerating the training speed by more than $10\times$.

\section{Our Approach}

\subsection{Baseline Setup}
We use a strong object detector $\mathcal{H}$-DETR+ResNet$50$ with AP=$52.2$ as our baseline for the following ablation experiments and report the results based on stronger $\mathcal{H}$-DETR+Swin-L and DINO-DETR+FocalNet-L for the system-level comparisons. The entire $\mathcal{H}$-DETR+ResNet$50$ model is pre-trained on Object$365$~\cite{shao2019objects365} and then fine-tuned on COCO for higher performance.

\begin{figure}[t]
\centering
\includegraphics[width=0.5\textwidth]{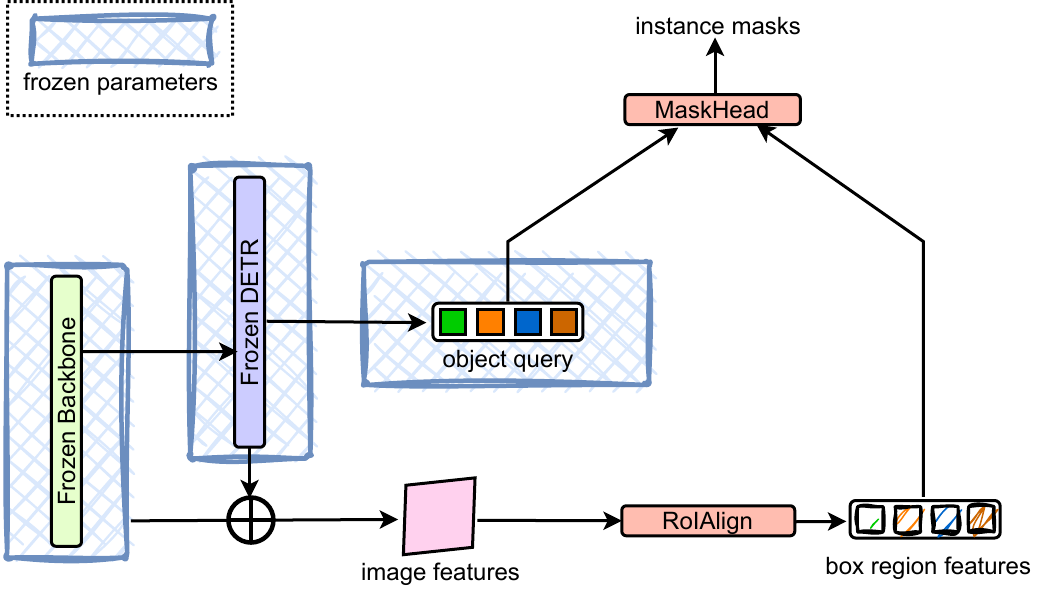}
\caption{\small{Mask Frozen-DETR Baseline: both RoIAlign and MaskHead are non-parametric operations.}}
\label{fig:frozen_detr_pipeline_0}
\end{figure}

To build a simple baseline for instance segmentation without extra training, we first sort the object queries output by the last Transformer decoder layer of the object detection model according to the decreasing order of their classification scores, and select the top $\sim$$100$ object queries for mask prediction. Then we multiply the object queries $\{\mathbf{q}_i|\mathbf{q}_i\in \mathbb{R}^{\mathsf{d}}\}_{i=1}^{N}$ and the image features $\mathbf{F}\in \mathbb{R}^{\frac{\mathsf{HW}}{16}\times\mathsf{d}}$ at $1$/$4$-resolution to get the instance segmentation masks as follows:
\begin{equation}
\begin{aligned}
\label{eq.mask1}
\mathbf{F} &= \mathbf{C}_1 + \operatorname{interpolate}(\mathbf{E}),\\
{\mathbf{M}_i} &= \operatorname{interpolate}(\operatorname{reshape}(\operatorname{Sigmoid}(\mathbf{q}_i\mathbf{F}^\top))),
\end{aligned}
\end{equation}
where $\mathbf{C}_1\in \mathbb{R}^{\frac{\mathsf{HW}}{16}\times\mathsf{d}}$ represents the feature map output by the first stage of the backbone. $\mathbf{E}\in \mathbb{R}^{\frac{\mathsf{HW}}{64}\times\mathsf{d}}$ represents the 1/8-resolution feature map from the Transformer encoder. 
$\mathsf{H}$, $\mathsf{W}$, and $\mathsf{d}$ represent the image height, image width, and feature hidden dimension, respectively. $\mathbf{M}_i \in \mathbb{R}^{\mathsf{HW}}$ represents the final predicted probability mask with the same resolution as the input image.
We illustrate the overall pipeline in Figure~\ref{fig:frozen_detr_pipeline_0}.

Next, we compute the confidence scores that reflects the quality of masks following:
\begin{equation}
\begin{aligned}
\label{eq.score}
{\mathrm{s}_i} = {\mathrm{c}_i} \times \frac{\operatorname{sum}(\mathbf{M}_i[\mathbf{M}_i > 0.5])}{\operatorname{sum}([\mathbf{M}_i > 0.5])},
\end{aligned}
\end{equation}
where $\mathrm{c}_i$ represents the classification score associated with the $i$-th object query predicted by the last Transformer decoder layer of the object detection model.

\begin{table}[t]
\begin{minipage}[t]{1\linewidth}
\vspace{2mm}
\centering
\setlength{\tabcolsep}{1pt}
\footnotesize
\renewcommand{\arraystretch}{1.2}
\resizebox{1.0\linewidth}{!}
{
\begin{tabular}{c|c|c|c|cccccc}
RoIAlign & 1/4 feat.  & \#FLOPs$\blacktriangle$ & \#params$\blacktriangle$ & AP$^{\rm{mask}}$ & AP$^{\rm{mask}}_{50}$ & AP$^{\rm{mask}}_{75}$ & AP$^{\rm{mask}}_{\rm{S}}$ & AP$^{\rm{mask}}_{\rm{M}}$ & AP$^{\rm{mask}}_{\rm{L}}$ \\
\shline
\xmark & \xmark  & $1.49$ G & $0.0$ M & $0.0$ & $0.1$ & $0.0$ & $0.0$ & $0.0$  & $0.1$ \\
\cmark & \xmark  & $0.53$ G & $0.0$ M & $4.4$ & $14.8$ & $1.3$ & $2.7$ & $4.6$ & $7.5$ \\
\xmark & \cmark  & $1.50$ G & $0.0$ M & $0.0$ & $0.1$ & $0.0$ & $0.0$ & $0.0$ & $0.1$ \\
\rowcolor{gray!10}\cmark & \cmark  & $0.54$ G & $0.0$ M & $5.1$ & $17.0$ & $1.5$ & $3.2$ & $4.9$  & $8.4$ \\
\end{tabular}
}
\caption{\small{\textbf{
Effect of each factor within our baseline that requires no training.}}
RoIAlign: use RoIAlign to pool the region features according to the predicted boxes.
1/4 feat.: fuse the $1$/$4$-resolution feature maps output by the stage-$1$ of backbone with the up-sampled $1$/$4$-resolution feature maps output by the Transformer encoder.
We use $\blacktriangle$ to mark the additional increased number of parameters and FLOPs in all the following tables.
}
\label{tab:baseline_ablate}
\end{minipage}
\end{table}

We illustrate the modifications when using RoIAlign operation~\cite{he2017mask} as follows. Instead of using the entire image features $\mathbf{F}\in \mathbb{R}^{\frac{\mathsf{HW}}{16}\times\mathsf{d}}$, we use the RoIAlign to gather the region feature maps located within the predicted bounding boxes:
\begin{equation}
\begin{aligned}
\label{eq.roi_mask}
{\mathbf{R}_i} = \operatorname{reshape}(\operatorname{RoIAlign}(\operatorname{reshape}(\mathbf{F}), \mathbf{b}_i)),
\end{aligned}
\end{equation}
where $\mathbf{R}_i\in \mathbb{R}^{\mathsf{hw}\times\mathsf{d}}$, $\mathbf{b}_i$ represents the predicted bounding box of $\mathbf{q}_i$.
We set $\mathsf{h}$ and $\mathsf{w}$ as $32$ by default.
Then we compute the instance segmentation masks as follows:
\begin{equation}
\begin{aligned}
\label{eq.paste_mask}
{\mathbf{M}^r_i} = \operatorname{paste}(\operatorname{interpolate}(\operatorname{reshape}(\operatorname{Sigmoid}(\mathbf{q}_i\mathbf{R}_{i}^\top)))),
\end{aligned}
\end{equation}
where we first reshape and interpolate the predicted regional instance masks to be the same size as the real bounding box size in the original image and then paste the resized ones to an empty instance mask of the same size as the original image.
We compute the confidence scores based on ${\mathbf{M}^r_i}$ following a similar manner.

\vspace{1mm}
\noindent\textbf{Results.}
Table~\ref{tab:baseline_ablate} shows the comparison results on the effect of fusing the $1$/$4$-resolution feature maps output by the first stage of the backbone and using the RoIAlign to constrain the computation focus only within the predicted bounding boxes.
According to the reported results,
we find that directly multiplying the object queries with the image feature maps performs very poorly, i.e., the best setting achieves a mask AP score $5.1\%$.
Notably, we also report the additional increased computational cost and number of parameters in all ablation experiments by default.

We aim to make minimal modifications to boost the segmentation performance based on the above baseline setting.
We show how to improve the system from three main aspects, including image feature encoder design, box region feature encoder design, and query feature encoder design, in the following discussions.
{We conduct all the following ablation experiments by freezing the entire object detection network and only fine-tuning the additional introduced parameters for $\sim$$6$ epochs on COCO instance segmentation task.}

\noindent\textbf{Experiment Setup.}
We use AdamW optimizer with an initial learning rate $1.5 \times 10^{-4}$, $\beta_1=0.9$, $\beta_2=0.999$ and a weight decay of $5 \times 10^{-5}$ is employed. We train the models for $88,500$ iterations (i.e. $6$ epochs), and divide the learning rate by $10$ at $0.9$ and $0.95$ fractions of the total number of the training iterations.
We follow the data pre-processing scheme of Deformable DETR \cite{zhu2020deformable}. We set the batch size as $8$ and run all experiments with V100 GPUs with $16$GB memory.
We report a set of COCO metrics including AP, AP$_{50}$, AP$_{75}$, AP$_{S}$, AP$_{M}$ and AP$_{L}$.

\begin{figure}[t]
\centering
\includegraphics[width=0.5\textwidth]{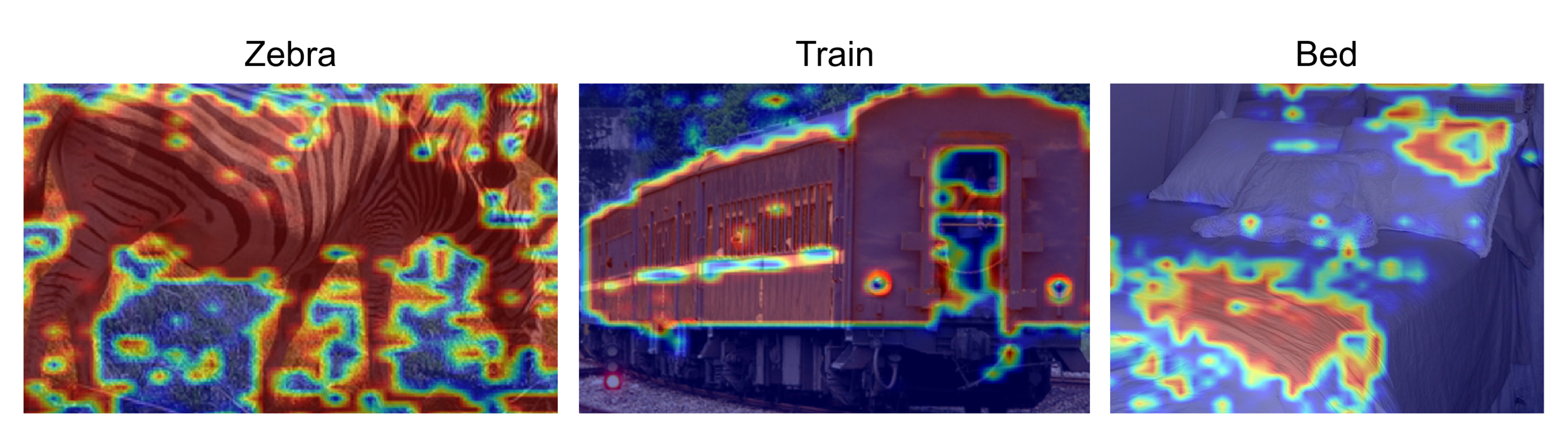}
\caption{\small{Coarse instance segmentation with the Mask frozen-DETR baseline. We visualize the predicted probability maps and the color indicates the confidence scores: red for high and blue for low.}}
\label{fig:baseline_seg}
\vspace{-3mm}
\end{figure}

\subsection{Image Feature Encoder}

\begin{figure}[t]
\centering
\includegraphics[width=0.5\textwidth]{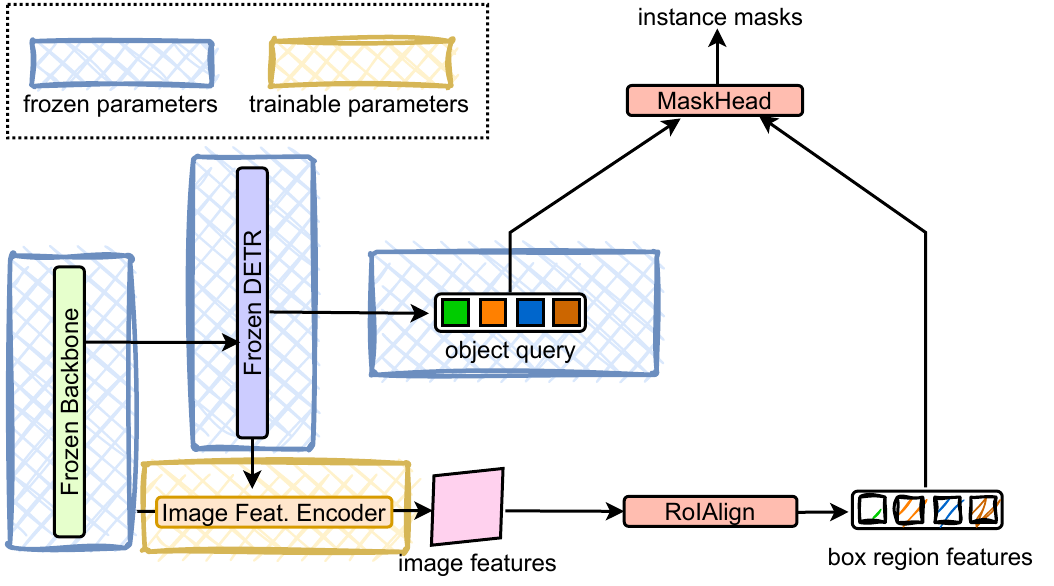}
\caption{\small{Add image feature encoder to Mask Frozen-DETR. We insert an additional image feature encoder to enhance the image feature maps.}}
\label{fig:frozen_detr_pipeline_2}
\end{figure}

We first study how to improve the previous baseline setting by introducing a trainable image feature encoder to transform the image feature maps into a more suitable feature space for instance segmentation tasks.
Figure~\ref{fig:frozen_detr_pipeline_2} shows the overall pipeline. We apply the image feature encoder on feature map $\mathbf{E}$ from the Transformer encoder. We simply modify Equation~\ref{eq.mask1} as follows:

\begin{equation}
\begin{aligned}
\label{eq.mask2}
\mathbf{F} &= \mathbf{C}_1 + \operatorname{interpolate}({{\mathcal{F}_e}(\mathbf{E})}),
\end{aligned}
\end{equation}
where the $\mathcal{F}_e(\cdot)$ represents the image feature encoder that refines the image feature map for all object queries simultaneously.
We study the following three kinds of modern convolution or transformer blocks.

\vspace{1mm}
\noindent\textbf{Deformable encoder block~\cite{zhu2020deformable}.}
We follow the multi-scale deformable encoder design and stack multiple multi-scale deformable encoder blocks to enhance the multi-scale feature map $\mathbf{E}$ following:
\begin{equation}
\begin{aligned}
\label{eq.deformable_image_encoder:1}
\mathbf{E} &= [\mathbf{E}_1, \mathbf{E}_2, \mathbf{E}_3, \mathbf{E}_4],\\
\mathcal{F}_e(\mathbf{E}) &= \operatorname{MultiScaleDeformableEnc}([\mathbf{E}_1, \mathbf{E}_2, \mathbf{E}_3, \mathbf{E}_4]),
\end{aligned}
\end{equation}
where $\mathbf{E}_1$, $\mathbf{E}_2$, $\mathbf{E}_3$, and $\mathbf{E}_4$ represent the feature maps of different scales from the Transformer encoder of the object detection system.
Each multi-scale deformable encoder block is implemented with $\operatorname{MSDeformAttn}\to\operatorname{LayerNorm}\to\operatorname{FFN}\to\operatorname{LayerNorm}$.
$\operatorname{FFN}$ is implemented as $\operatorname{Linear}\to\operatorname{GELU}\to\operatorname{Linear}$ by default.

\vspace{1mm}
\noindent\textbf{Swin Transformer encoder block~\cite{liu2021swin}}
We follow the Swin Transformer to apply a stack of multiple Swin Transformer blocks on the feature map $\mathbf{E}_1$ with highest resolution as follows:
\begin{equation}
\begin{aligned}
\label{eq.swin_image_encoder:1}
\mathcal{F}_e(\mathbf{E}_1) &= \operatorname{SwinTransformerEnc}(\mathbf{E}_1),
\end{aligned}
\end{equation}
where each Swin Transformer block is implemented as $\operatorname{LayerNorm}\to\operatorname{W-MSA}\to\operatorname{LayerNorm}\to\operatorname{FFN}$. $\operatorname{W-MSA}$ represents the window multi-head self-attention operation.
We apply shifted $\operatorname{W-MSA}$ in the successive block to propogate information across windows following~\cite{liu2021swin}.

\vspace{1mm}
\noindent\textbf{ConvNext encoder block~\cite{liu2022convnet}}
We follow the ConvNext to apply the proposed combination of large-kernel convolution and inverted bottleneck on the Transformer encoder feature map $\mathbf{E}_1$ following:
\begin{equation}
\begin{aligned}
\label{eq.deformable_image_encoder:1}
\mathcal{F}_e(\mathbf{E}_1) &= \operatorname{ConvNextBlock}(\mathbf{E}_1).
\end{aligned}
\end{equation}
where each ConvNext block is implemented as $\operatorname{DWC}\to\operatorname{LayerNorm}\to\operatorname{FFN}$. $\operatorname{DWC}$ represents a depth-wise convolution with large kernel size, i.e., $7\times7$.

\begin{table}[t]
\begin{minipage}[t]{1\linewidth}
\vspace{2mm}
\centering
\setlength{\tabcolsep}{2pt}
\footnotesize
\renewcommand{\arraystretch}{1.2}
\resizebox{1.0\linewidth}{!}
{
\begin{tabular}{l|c|c|c|cccccc}
block type & \# layers  & \#FLOPs$\blacktriangle$ & \#params$\blacktriangle$ & AP$^{\rm{mask}}$ & AP$^{\rm{mask}}_{50}$ & AP$^{\rm{mask}}_{75}$ & AP$^{\rm{mask}}_{\rm{S}}$ & AP$^{\rm{mask}}_{\rm{M}}$ & AP$^{\rm{mask}}_{\rm{L}}$ \\
\shline
None & $0$  & $0.54$ G & $0.0$ M & $5.1$ & $17.0$ & $1.5$ & $3.2$ & $4.9$  & $8.4$ \\\hline
deformable. & $1$  & $14.71$ G & $0.76$ M & $30.8$ & $58.2$ & $29.5$ & $13.5$ & $32.7$ & $49.0$  \\
\rowcolor{gray!10}deformable. & $2$  & $28.88$ G & $1.51$ M & $32.9$ & $60.0$ & $32.4$ & $14.9$ & $35.3$ & $52.1$  \\
deformable. & $3$  & $43.04$ G & $2.27$ M & $33.7$ & $60.5$ & $33.7$ & $15.1$ & $36.3$ & $53.4$ \\\hline
Swin Trans. & $1$   & $14.20$ G & $0.80$ M & $30.0$ & $58.0$ & $28.0$ & $13.8$ & $32.3$ & $46.3$  \\
Swin Trans. & $2$   & $27.86$ G & $1.59$ M & $31.6$ & $59.2$ & $30.4$ & $14.5$ & $34.2$ & $48.6$ \\
Swin Trans. & $3$   & $41.51$ G & $2.39$ M &$32.2$ & $59.6$ & $31.3$ & $14.9$ & $34.8$ & $49.5$ \\\hline
ConvNext. & $1$  & $8.12$ G & $0.54$ M & $27.5$ & $55.9$ & $24.4$ & $12.7$ & $29.8$ & $42.1$   \\
ConvNext. & $2$  & $15.70$ G & $1.08$ M & $30.0$ & $58.1$ & $27.8$ & $13.9$ & $32.4$ & $46.0$  \\
ConvNext. & $3$  & $23.27$ G & $1.62$ M & $30.7$ & $58.5$ & $28.9$ & $14.3$ & $33.2$ & $46.9$ \\
\end{tabular}
}
\caption{\small{{
Effect of image feature encoder design.}}
}
\label{tab:feat_enc_ablate}
\end{minipage}
\end{table}

\vspace{1mm}
\noindent\textbf{Results.}
In Table~\ref{tab:feat_enc_ablate}, we compare different choices of the image feature encoder architecture design. We observe that: (i) All three image feature encoder implementations improve the instance segmentation performance. (ii) More image feature encoder layers leads to better performance, e.g., $3\times$ deformable encoder layers: AP=$32.9\%$ vs. $1\times$ deformable encoder layer: AP=$30.8\%$. (iii) Under similar computation budget, Deformable encoder block performs better, e.g., $2\times$ deformable encoder layers: AP=$32.9\%$/FLOPs=$28.88$G vs. $2\times$ Swin transformer encoder layers: AP=$31.6\%$/FLOPs=$27.86$G vs. $3\times$ ConvNext encoder layers: AP=$30.7\%$/FLOPs=$23.27$G. We use $2\times$ deformable encoder layers as the image feature encoder in the following experiments as it has a better trade-off between performance and computational cost.

\subsection{Box Feature Encoder}
\begin{figure}[t]
\centering
\includegraphics[width=0.5\textwidth]{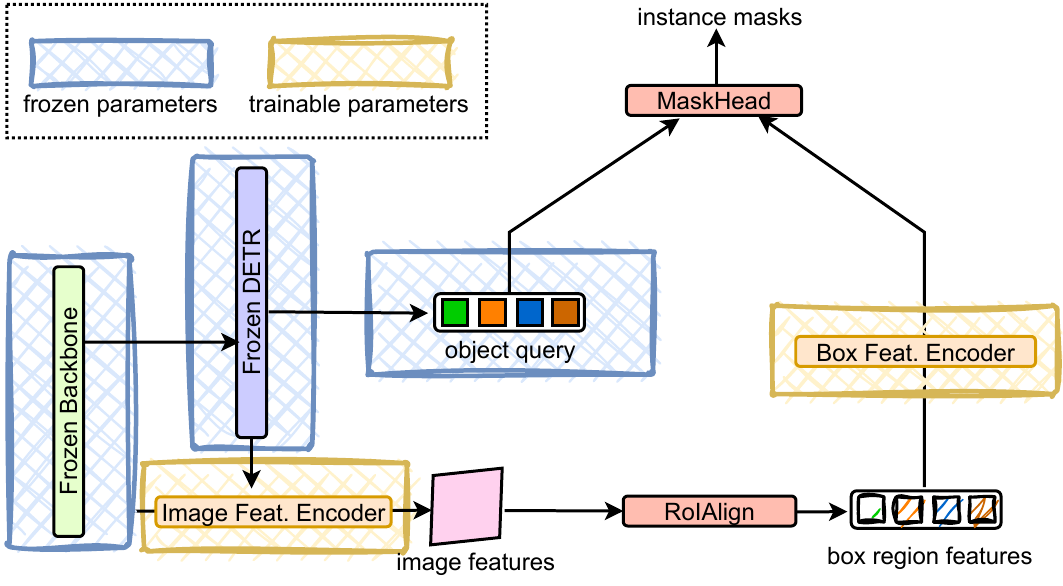}
\caption{\small{Add box feature encoder to Mask Frozen-DETR. We insert a feature encoder to enhance the bounding box region features.}}
\label{fig:frozen_detr_pipeline_3}
\end{figure}

Now we study the influence of improving the box region features with an additional box feature encoder design.
We illustrate the modification in Figure~\ref{fig:frozen_detr_pipeline_3}.
we simply apply additional transformation $\mathcal{F}_b$ on ${\mathbf{R}_i}$ before Equation~\ref{eq.paste_mask} following:
\begin{equation}
\begin{aligned}
\label{eq.roi_mask}
{\mathbf{R}_i} = {\mathcal{F}_b}({\mathbf{R}_i}),
\end{aligned}
\end{equation}
where we study different choices of the box feature encoder implementation following the study on the image feature encoder design.

\begin{table}[t]
\begin{minipage}[t]{1\linewidth}
\vspace{2mm}
\centering
\setlength{\tabcolsep}{1.2pt}
\footnotesize
\renewcommand{\arraystretch}{1.2}
\resizebox{1.0\linewidth}{!}
{
\begin{tabular}{l|c|c|c|cccccc}
block type & \# layers & \#FLOPs$\blacktriangle$ & \#params$\blacktriangle$ & AP$^{\rm{mask}}$ & AP$^{\rm{mask}}_{50}$ & AP$^{\rm{mask}}_{75}$ & AP$^{\rm{mask}}_{\rm{S}}$ & AP$^{\rm{mask}}_{\rm{M}}$ & AP$^{\rm{mask}}_{\rm{L}}$  \\
\shline
None & $0$  & $28.88$ G & $1.51$ M & $32.9$ & $60.0$ & $32.4$ & $14.9$ & $35.3$ & $52.1$ \\\hline
deformable. & $1$  & $98.55$ G & $2.46$ M & $44.4$ & $67.6$ & $48.0$ & $24.5$ & $47.7$ & $62.8$ \\
\rowcolor{gray!10}deformable. & $2$  & $168.23$ G & $3.14$ M & $44.9$ & $67.7$ & $48.6$ & $24.9$ & $48.2$ & $63.4$ \\
deformable. & $3$  &$237.91$ G & $3.82$ M & $45.1$ & $67.8$ & $48.9$ & $25.2$ & $48.4$ & $63.7$ \\\hline
Swin Trans. & $1$   & $112.84$ G & $2.31$ M & $42.5$ & $66.3$ & $45.4$ & $23.0$ & $45.5$ & $61.2$ \\
Swin Trans. & $2$   & $196.81$ G & $3.10$ M & $43.7$ & $66.9$ & $46.8$ & $23.8$ & $46.8$ & $62.3$ \\
Swin Trans. & $3$   & $280.77$ G & $3.89$ M & $44.3$ & $67.3$ & $47.6$ & $24.4$ & $47.6$ & $62.8$ \\\hline
ConvNext. & $1$  & $83.90$ G & $2.05$ M & $40.9$ & $65.8$ & $43.4$ & $21.4$ & $43.8$ & $59.9$   \\
ConvNext. & $2$  & $138.92$ G & $2.59$ M & $43.0$ & $66.9$ & $46.0$ & $23.4$ & $46.1$ & $61.7$  \\
ConvNext. & $3$  & $193.95$ G & $3.13$ M & $43.7$ & $67.2$ & $47.1$ & $23.9$ & $47.0$ & $62.1$  \\
\end{tabular}
}
\caption{\small{{
Effect of box feature encoder design.}}
}
\label{tab:box_enc_ablate}
\end{minipage}
\end{table}

\begin{table}[t]
\begin{minipage}[t]{1\linewidth}
\vspace{2mm}
\centering
\setlength{\tabcolsep}{2pt}
\footnotesize
\renewcommand{\arraystretch}{1.2}
\resizebox{1.0\linewidth}{!}
{
\begin{tabular}{c|c|c|cccccc}
\# hidden dim. & \#FLOPs$\blacktriangle$ & \#params$\blacktriangle$ & AP$^{\rm{mask}}$ & AP$^{\rm{mask}}_{50}$ & AP$^{\rm{mask}}_{75}$ & AP$^{\rm{mask}}_{\rm{S}}$ & AP$^{\rm{mask}}_{\rm{M}}$ & AP$^{\rm{mask}}_{\rm{L}}$  \\
\shline
$256$ & $168.23$ G & $3.14$ M & $44.9$ & $67.7$ & $48.6$ & $24.9$ & $48.2$ & $63.4$ \\
\rowcolor{gray!10}$128$ & $66.60$ G & $2.07$ M & $44.7$ & $67.6$ & $48.2$ & $24.8$ & $48.1$ & $63.2$ \\
$96$ & $50.76$ G & $1.87$ M & $44.6$ & $67.5$ & $48.2$ & $24.7$ & $47.9$ & $63.2$ \\
$64$ & $39.11$ G & $1.71$ M & $44.5$ & $67.5$ & $48.0$ & $24.7$ & $47.9$ & $63.0$ \\
\end{tabular}
}
\caption{\small{{
Effect of hidden dimension after channel mapper.}}
}
\label{tab:box_enc_ablate_channel_deformable}
\end{minipage}
\end{table}

\vspace{1mm}
\noindent\textbf{Channel mapper.}
To build an efficient box feature encoder, we propose to use a channel mapper as a simple $\operatorname{Linear}$ layer to decrease the channel dimension of $\mathbf{F}\in\mathbb{R}^{\frac{\mathsf{HW}}{16}\times\mathsf{d}}$ and apply the box region feature encoder on the updated feature map with smaller channels.

\vspace{1mm}
\noindent\textbf{Results.}
We compare the effect of box feature encoder architecture design in Table~\ref{tab:box_enc_ablate}. We notice significant computational cost increase for all settings, mainly due to two reasons: (i) the large number of bounding box region features, i.e., $100$ box predictions during inference; and (ii) the high-resolution of the bounding box feature map, i.e., $32\times32$. We study the impact of different resolutions of the region feature maps in the ablation experiments. We find similar conclusions as Table~\ref{tab:feat_enc_ablate}: all methods improve the performance, more encoder layers lead to better performance, and deformable encoder block performs the best. Therefore, we use deformable encoder blocks for the box feature encoder. To reduce the expensive computational cost, we use the channel mapper to decrease the hidden dimension. The results are in Table~\ref{tab:box_enc_ablate_channel_deformable}. We observe that lower hidden dimension reduces computational cost with little performance loss. Considering the trade-off between performance and overhead, we use $128$ hidden dimensions for subsequent experiments.

\subsection{Query Feature Encoder}
After studying the influence of adding the image feature encoder and box region feature encoder, we further investigate how to design a suitable query feature encoder to refine the object queries originally designed for detecting the boxes for instance segmentation tasks.

\vspace{1mm}
\noindent\textbf{Object-to-object attention.}
The proximity of objects to each other may results in a situation where multiple instances lie within one bounding box. We add object-to-object attention to help object query representations distinguish instances. Specifically, we use multi-head self-attention mechanism to process the queries as follows:
\begin{equation}
\begin{aligned}
\label{eq.object2object_attn}
[\mathbf{q}_1, \mathbf{q}_2, \cdots, \mathbf{q}_N] = \operatorname{SelfAttention}([\mathbf{q}_1, \mathbf{q}_2, \cdots, \mathbf{q}_N]).
\end{aligned}
\end{equation}

\vspace{1mm}
\noindent\textbf{Box-to-object attention.}
In the frozen DETR-based object detector, the object queries are used to perform object detection and process the whole image feature instead of a box region feature. The discrepancies between the usage of object queries in the frozen detector and MaskHead may lead to sub-optimal segmentation results. Therefore, we introduce box-to-object attention to transform the queries and adapt them to the segmentation task as follows:
\begin{equation}
\begin{aligned}
\label{eq.box2feat_attn}
{\mathbf{q}_i} = \operatorname{CrossAttention}(\mathbf{q}_i, \mathbf{R}_i),
\end{aligned}
\end{equation}
where the object queries are updated by refering to the information in the box region feature.

\vspace{1mm}
\noindent\textbf{FFN.}
The feed forward network (FFN) block is a widely employed element in Transformers. It is usually integrated after an attention layer to transform individual tokens. In Table \ref{tab:query_enc_ablate}, we investigate the efficacy of this block in adjusting object query representations for segmentation. 

\vspace{1mm}
\noindent\textbf{Results.}
Table~\ref{tab:query_enc_ablate} shows the comparison results on the effect of different modifications to the query feature encoder architecture design.
We find no performance improvement when only using FFN to enhance the object queries and a minor gain (+$0.1$) when only using the box-to-object attention module. We think this is because the enhanced box features are already strong for instance segmentation, so using the original object query representation from the frozen detection model is enough. Moreover, using object-to-object attention reduces the performance, as the interaction between object queries might mix the semantic information of different objects. In general, we conclude that transforming the object queries is unnecessary and simply using the original ones to interact with the refined box region features achieves strong results.
We only adopt the box-to-object attention design in the qualitative analysis experiments.

\begin{figure}[t]
\centering
\includegraphics[width=0.5\textwidth]{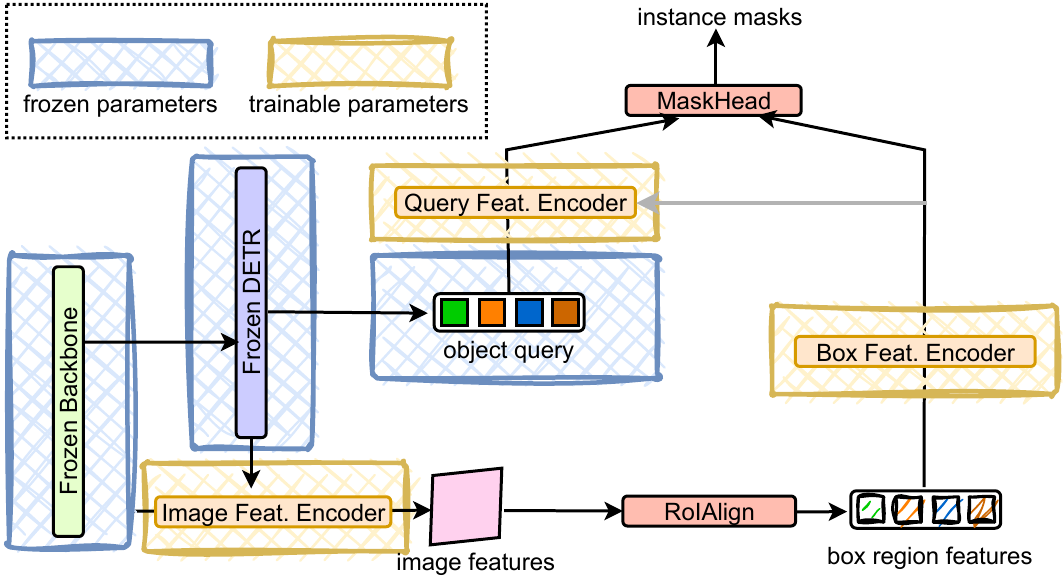}
\caption{\small{Add object query encoder to Frozen DETR.} We apply a query feature encoder to enhance the object query representations. This is the complete framework of our Mask Frozen-DETR.}
\label{fig:frozen_detr_pipeline}
\end{figure}

\begin{table}[t]
\begin{minipage}[t]{1\linewidth}
\vspace{2mm}
\centering
\setlength{\tabcolsep}{1.5pt}
\footnotesize
\renewcommand{\arraystretch}{1.2}
\resizebox{1.0\linewidth}{!}
{
\begin{tabular}{c|c|c|c|c|cccccc}
FFN & O2O & B2O  & \#FLOPs$\blacktriangle$ & \#params$\blacktriangle$ & AP$^{\rm{mask}}$ & AP$^{\rm{mask}}_{50}$ & AP$^{\rm{mask}}_{75}$ & AP$^{\rm{mask}}_{\rm{S}}$ & AP$^{\rm{mask}}_{\rm{M}}$ & AP$^{\rm{mask}}_{\rm{L}}$ \\
\shline
\xmark & \xmark & \xmark  & $66.60$ G & $2.07$ M & $44.7$ & $67.6$ & $48.2$ & $24.8$ & $48.1$ & $63.2$ \\
\cmark & \xmark & \xmark  & $66.63$ G & $2.33$ M & $44.7$ & $67.6$ & $48.3$ & $24.9$ & $48.0$ & $63.2$ \\
\cmark & \cmark & \xmark  & $66.65$ G & $2.53$ M & $43.9$ & $67.1$ & $47.3$ & $24.4$ & $47.3$ & $61.8$ \\
\rowcolor{gray!10}\cmark  & \xmark & \cmark  & $73.40$ G & $2.46$ M & $44.8$ & $67.6$ & $48.4$ & $24.9$ & $48.1$ & $63.4$ \\
\end{tabular}
}
\caption{\small{{
Effect of each factor within the query feature encoder.}
O2O: object-to-object attention module.
B2O: box-to-object attention module.
}
}
\label{tab:query_enc_ablate}
\end{minipage}
\end{table}

\subsection{Other Improvements}

\vspace{1mm}
\noindent\textbf{Mask loss on sampled pixel points.}
Inspired by implicit PointRend~\cite{cheng2022pointly} that shows a segmentation model can be trained with its mask loss computed on $N$ sampled points instead of the entire mask, we compute the mask loss with sampled points in the  the final loss calculation. Specifically, 
given number of points $N$, oversample ratio $k$ ($k > 1)$ and importance sample ratio $\beta$ ($\beta \in [0, 1]$), we randomly sample $kN$ points from the output mask and select $\beta N$ most uncertain points from the sampled points. Then we randomly sample other $(1-\beta)N$ points from the output mask and compute loss only on these $N$ points.

\begin{table}[t]
\begin{minipage}[t]{1\linewidth}
\vspace{2mm}
\centering
\setlength{\tabcolsep}{0.5pt}
\footnotesize
\renewcommand{\arraystretch}{1.2}
\resizebox{1.0\linewidth}{!}
{
\begin{tabular}{ccc|c|c|cccccc}
   neck & \thead{samp. pixel \\ sup.} & \thead{mask \\ scoring} & \#FLOPs$\blacktriangle$ & \#params$\blacktriangle$ & AP$^{\rm{mask}}$ & AP$^{\rm{mask}}_{50}$ & AP$^{\rm{mask}}_{75}$ & AP$^{\rm{mask}}_{\rm{S}}$ & AP$^{\rm{mask}}_{\rm{M}}$ & AP$^{\rm{mask}}_{\rm{L}}$  \\
    \shline
   \xmark & \xmark & \xmark & $73.40$ G & $2.46$ M & $44.8$ & $67.6$ & $48.4$ & $24.9$ & $48.1$ & $63.4$ \\
   \cmark & \xmark & \xmark & $77.07$ G & $2.53$ M & $45.2$ & $67.8$ & $48.9$ & $25.3$ & $48.5$ & $63.8$ \\
   \cmark & \cmark & \xmark & $77.07$ G & $2.53$ M & $45.3$ & $67.8$ & $49.2$ & $25.4$ & $48.6$ & $64.1$ \\ 
  \rowcolor{gray!10}\cmark & \cmark & \cmark & $82.09$ G& $3.26$ M & $45.7$ & $67.5$ & $49.8$ & $25.6$ & $48.9$ & $64.1$ \\
\end{tabular}
}
\caption{\small{{
Effect of other improvements including sampled pixel supervision, mask scoring, and neck design.}}
}
\label{tab:other_improvement_ablate}
\end{minipage}
\end{table}

\begin{table*}[t]
\begin{minipage}[t]{1\linewidth}
\centering\setlength{\tabcolsep}{5pt}
\footnotesize
\renewcommand{\arraystretch}{1.2}
\label{tab:coco_instance_exp}
\resizebox{\linewidth}{!}
{
\begin{tabular}{l|l|c|c|c|c|cccccc}
method & backbone & \#epochs  & Object$365$ & AP$^{\rm{box}}$ & AP$^{\rm{mask}}$ & AP$^{\rm{mask}}_{50}$ & AP$^{\rm{mask}}_{75}$ & AP$^{\rm{mask}}_{\rm{S}}$ & AP$^{\rm{mask}}_{\rm{M}}$ & AP$^{\rm{mask}}_{\rm{L}}$ & GPU Hours \\
\shline
K-Net-N$256$~\cite{zhang2021k} & R$50$ & $36$ & \xmark & $-$ & $38.6$ & $60.9$ & $41.0$ & $19.1$ & $42.0$ & $57.7$ & $-$\\
QueryInst~\cite{FangQueryInst} & Swin-L & $50$ & \xmark & $56.1$ & $48.9$ & $74.0$ & $53.9$ & $30.8$ & $52.6$ & $68.3$ & $-$ \\

Mask$2$Former~\cite{cheng2021masked} & R$50$ & $50$ & \xmark & $-$ & $43.7$ & $-$ & $-$ & $23.4$ & $47.2$ & $64.8$ &  $502$  \\
Mask$2$Former~\cite{cheng2021masked} & Swin-L & $100$ & \xmark & $-$ & ${50.1}$ & $-$ & $-$ & $29.9$ & $53.9$ & $72.1$ &  $1,700$  \\ 
Mask DINO~\cite{li2022mask} & R$50$ & $50$ & \xmark & $50.5$ & $46.0$ & $68.9$ & $50.3$ & $26.0$ & $49.3$ & $65.5$ & $1,404$ \\
Mask DINO~\cite{li2022mask} & Swin-L & $50$ & \xmark & $58.3$ & $52.1$ & $76.5$ & $57.6$ & $32.9$ & $55.4$ & $72.5$ & $2,400$ \\
\rowcolor{gray!8}Mask Frozen-$\mathcal{H}$-DETR &R$50$ & $6$ & \xmark& $49.9$ & $44.1$ & $66.2$ & $47.8$ & $24.4$ & $47.0$ & $62.6$ & $49$ \\
\rowcolor{gray!8}Mask Frozen-$\mathcal{H}$-DETR & Swin-L & $6$ & \xmark & $59.1$ & $51.9$ & $75.8$ & $57.2$ & $31.6$ & $55.1$ & $71.6$ & $179$\\
\hline
ViT-Adapter-L~\cite{chen2022vitadapter} & ViT-L & $8$ & \cmark & $61.8$ & $53.0$ & $-$ & $-$ & $-$ & $-$ & $-$ & $1,068$ \\
Mask DINO~\cite{li2022mask} & Swin-L & $24$ & \cmark & $-$ & $54.5$ & $-$ & $-$ & $-$ & $-$ & $-$ & $1,600$ \\
\rowcolor{gray!8}Mask Frozen-$\mathcal{H}$-DETR &R$50$ & $6$ & \cmark & $52.2$ & $45.7$ & $67.5$ & $49.8$ & $25.6$ & $48.9$ & $64.1$ & $49$\\
\rowcolor{gray!8}Mask Frozen-$\mathcal{H}$-DETR & Swin-L & $6$ & \cmark & $62.3$ & $54.0$ & $77.9$ & $59.5$ & $35.6$ & $57.4$ & $73.0$ & $172$\\
\rowcolor{gray!8}Mask Frozen-DINO-DETR & FocalNet-L & $6$ & \cmark& $63.2$ & $\bf{54.9}$ & $\bf{78.9}$ & $\bf{60.8}$ & $\bf{37.2}$ & $\bf{58.4}$ & $\bf{72.9}$ & ${136}$\\
\end{tabular}
}
\caption{\small{Comparison with SOTA instance segmentation methods on COCO val.}
}
\label{coco_instance_exp}
\vspace{2mm}
\end{minipage}
\begin{minipage}[t]{1\linewidth}
\vspace{2mm}
\centering
\setlength{\tabcolsep}{7pt}
\footnotesize
\renewcommand{\arraystretch}{1.35}
\resizebox{1.0\linewidth}{!}
{
\begin{tabular}{l|l|c|c|c|cccccc}
method & backbone & \#epochs  & Object$365$ & AP$^{\rm{box}}$ & AP$^{\rm{mask}}$ & AP$^{\rm{mask}}_{50}$ & AP$^{\rm{mask}}_{75}$ & AP$^{\rm{mask}}_{\rm{S}}$ & AP$^{\rm{mask}}_{\rm{M}}$ & AP$^{\rm{mask}}_{\rm{L}}$ \\
\shline
K-Net-N$256$~\cite{zhang2021k} & R$101$ & $36$  & \xmark & - & $40.6$ & $63.3$ & $43.7$ & $18.8$ & $43.3$ & $59.0$ \\
SOLQ~\cite{dong2021solq} & Swin-L & $50$ & \xmark & $56.5$ & $46.7$ & - & - & $29.2$ & $50.1$ & $60.9$ \\
SOIT~\cite{yu2022soit} & Swin-L & $36$ & \xmark &$56.9$ & $49.2$ & $74.3$ & $53.5$ & $30.2$ & $52.7$ & $65.2$ \\
QueryInst~\cite{FangQueryInst} & Swin-L & $50$ & \xmark & $56.1$ & $49.1$ & $74.2$ & $53.8$ & $31.5$ & $51.8$ & $63.2$ \\
Mask$2$Former~\cite{cheng2021masked} & Swin-L & $100$ & \xmark & - & $50.5$ & $74.9$ & $54.9$ & $29.1$ & $53.8$ & $\bf{71.2}$ \\
Mask DINO~\cite{li2022mask} & Swin-L & $24$ & \cmark & - & $54.7$ & - & - & - & - & - \\
\rowcolor{gray!8}Mask Frozen-DINO-DETR & FocalNet-L & $6$ & \cmark & $63.2$ & $\bf{55.3}$ & $\bf{79.3}$ & $\bf{61.4}$ & $\bf{37.8}$ & $\bf{58.4}$ & $70.4$ \\
\end{tabular}
}
\caption{\small{{
Comparison with SOTA instance segmentation methods on COCO test-dev.}}
}
\label{tab:coco_test}
\end{minipage}
\end{table*}

\vspace{1mm}
\noindent\textbf{Mask scoring.}
Since the classification scores predicted by the frozen DETR cannot reflect the quality of segmentation masks, we introduce mask scoring \cite{huang2019mask} to our method to adjust the score, making it able to describe the quality of segmentation masks more precisely. Specifically, the mask scoring head takes the output mask and box region features as input and uses them to predict the iou score between the output mask and ground truth following:
\begin{equation}
\begin{aligned}
\label{eq.mask_scoring}
{\mathrm{iou}_i} = \operatorname{MLP}(\operatorname{Flatten}(\operatorname{Conv}(\operatorname{Cat}(\mathbf{M}_i, \mathbf{R}_i)))),
\end{aligned}
\end{equation}
where $\operatorname{Cat}$, $\operatorname{Conv}$ and $\operatorname{MLP}$ refer to concatenation, covolution layers and multi layer perceptron, respectively.
The iou score predicted by the mask scoring head is then used to adjust the classification score as following:
\begin{equation}
\begin{aligned}
\label{eq.score}
{\mathrm{s}_i} = {\mathrm{c}_i} \mathrm{iou}_i,
\end{aligned}
\end{equation}
where $\mathrm{s}_i$ is the confidence score of the output mask.

\vspace{1mm}
\noindent\textbf{Neck for backbone feature.}
The feature map output by the first stage of backbone $\mathbf{C}_1\in \mathbb{R}^{\frac{\mathsf{HW}}{16}\times\mathsf{d}}$  may not contain sufficient semantic information for accurate instance segmentation. Therefore, we introduce a simple neck block to encode more semantic information into the high resolution feature map  $\mathbf{C}_1$. The neck block can be described using the following formula:
\begin{equation}
\begin{aligned}
\label{eq.mask_scoring}
{\mathbf{C}_1} = \operatorname{GN}(\operatorname{PWConv}(\mathbf{C}_1)),
\end{aligned}
\end{equation}
where $\operatorname{GN}$ and $\operatorname{PWConv}$ refer to group normalization and point-wise convolution, respectively.

\vspace{1mm}
\noindent\textbf{Results.}
In Table \ref{tab:other_improvement_ablate}, we attempt to further improve the results by integrating a neck block design, using sampled pixel supervision, and using mask scoring.
We observe that all three designs bring consistent gains in AP scores. For example, using the neck block improves AP from $44.8\%$ to $45.2\%$ and using mask scoring improves AP from $45.3\%$ to $45.7\%$. Notably, while sampled pixel supervision reduces the number of the points for training supervisions by $75\%$, it still brings a slight gain in AP (+$0.1\%$). Therefore, we use all three designs in the following experiments.

\section{Comparison with SOTA Systems}
To compare our system with state-of-the-art instance segmentation methods, we construct a series of strong Mask Frozen-DETR models based on different Frozen DETR-based detector weights. These include Mask Frozen-$\mathcal{H}$-DETR + ResNet-$50$, which uses $\mathcal{H}$-DETR + ResNet-$50$ with AP$^{\rm{box}}$ of $52.2\%$; Mask Frozen-$\mathcal{H}$-DETR + Swin-L, which uses $\mathcal{H}$-DETR + Swin-L with AP$^{\rm{box}}$ of $62.3\%$; and Mask Frozen-DINO-DETR + Swin-L, which uses DINO-DETR + FocalNet-L with AP$^{\rm{box}}$ of $63.2\%$.

Table~\ref{coco_instance_exp} presents the detailed comparison results on COCO \texttt{val} set. We can see that, with Object$365$ object detection pre-training, our approach surpasses the very recent state-of-the-art Mask-DINO by a clear margin (ours: $54.9\%$ vs. Mask DINO: $54.5\%$). This result is remarkable considering that the strong object detector DINO-DETR achieves even better object detection performance, i.e., $63.2\%$, than the DINO-DETR + FocalNet-L that we use. The most important advantage of our approach is the significantly reduced training time, e.g., we can complete the training of DINO-DETR + FocalNet-L within \textbf{$17\times$ hours} while training a Mask DINO + Swin-L takes more than \textbf{$8\times$ days} when using $8\times$ V100 GPUs.

We also provide the detailed comparison results for Mask Frozen-$\mathcal{H}$-DETR + ResNet-$50$ and Mask Frozen-$\mathcal{H}$-DETR + Swin-L. In general, our approach achieves competitive results across various model sizes and different DETR-based frameworks.
We further compare Mask Frozen-DINO-DETR to the state-of-the-art methods in instance segmentation on COCO test-dev in Table \ref{tab:coco_test}. Frozen-DINO-DETR with FocalNet-L achieves an AP of $55.3\%$. This result surpasses the recent state-of-the-art method, Mask DINO, by +$0.6$ AP.

\section{Ablation Experiments and Analysis}

\begin{table}[t]
\begin{minipage}[t]{1\linewidth}
\vspace{2mm}
\centering
\setlength{\tabcolsep}{3pt}
\footnotesize
\renewcommand{\arraystretch}{1.2}
\resizebox{1.0\linewidth}{!}
{
\begin{tabular}{l|c|c|cccccc}
   output size & \#FLOPs$\blacktriangle$ & \#params$\blacktriangle$ & AP$^{\rm{mask}}$ & AP$^{\rm{mask}}_{50}$ & AP$^{\rm{mask}}_{75}$ & AP$^{\rm{mask}}_{\rm{S}}$ & AP$^{\rm{mask}}_{\rm{M}}$ & AP$^{\rm{mask}}_{\rm{L}}$  \\
    \shline
   $16 \times 16$ & $46.09$ G & $2.97$ M & $44.8$ & $67.5$ & $49.1$ & $25.4$ & $48.0$ & $62.0$\\ 
   \rowcolor{gray!10}$32 \times 32$ & $82.09$ G & $3.26$ M & $45.7$ & $67.5$ & $49.8$ & $25.6$ & $48.9$ & $64.1$ \\
   $64 \times 64$ & $225.86$ G & $4.44$ M & $45.9$ & $67.6$ & $50.2$ & $25.8$ & $49.1$ & $64.4$ \\
\end{tabular}
}
\caption{\small{{
Effect of RoIAlign output size.}}
}
\label{tab:roi_output_size}
\end{minipage}
\end{table}

\begin{table}[t]
\begin{minipage}[t]{1\linewidth}
\vspace{2mm}
\centering
\setlength{\tabcolsep}{9pt}
\footnotesize
\renewcommand{\arraystretch}{1.2}
\resizebox{1.0\linewidth}{!}
{
\begin{tabular}{c|cccccc}
   epoch  & AP$^{\rm{mask}}$ & AP$^{\rm{mask}}_{50}$ & AP$^{\rm{mask}}_{75}$ & AP$^{\rm{mask}}_{\rm{S}}$ & AP$^{\rm{mask}}_{\rm{M}}$ & AP$^{\rm{mask}}_{\rm{L}}$  \\
    \shline
   \rowcolor{gray!10}$6$ & $45.7$ & $67.5$ & $49.8$ & $25.6$ & $48.9$ & $64.1$ \\
   $12$  & $46.0$ & $67.7$ & $50.3$ & $25.9$ & $49.1$ & $64.3$ \\ 
\end{tabular}
}
\caption{\small{{
Effect of training epochs.}}
}
\label{tab:effect_training_epochs}
\end{minipage}
\end{table}

\begin{table}[t]
\begin{minipage}[t]{1\linewidth}
\vspace{2mm}
\centering
\setlength{\tabcolsep}{7pt}
\footnotesize
\renewcommand{\arraystretch}{1.2}
\resizebox{1.0\linewidth}{!}
{
\begin{tabular}{c|c|cccccc}
   LSJ   & GPU Hour& AP$^{\rm{mask}}$ & AP$^{\rm{mask}}_{50}$ & AP$^{\rm{mask}}_{75}$ & AP$^{\rm{mask}}_{\rm{S}}$ & AP$^{\rm{mask}}_{\rm{M}}$ & AP$^{\rm{mask}}_{\rm{L}}$  \\
    \shline
   \rowcolor{gray!10}\xmark & $49$ & $45.7$ & $67.5$ & $49.8$ & $25.6$ & $48.9$ & $64.1$ \\
  \cmark  & $63$ & $46.0$ & $67.7$ & $50.1$ & $26.1$ & $49.1$ & $64.0$ \\
\end{tabular}
}
\caption{\small{{
Effect of large scale jittering.}}
}
\label{tab:effect_lsj}
\end{minipage}
\end{table}

\vspace{1mm}
\noindent\textbf{RoIAlign output size.}
Table \ref{tab:roi_output_size} shows the the influence of RoIAlign output size. We observe that (i) Enlarging the RoIAlign output size significantly increases GFLOPs. (ii) Increasing RoIAlign output size can improve instance segmentation performance. Specifically, we observe a 0.9 improvement in AP by increasing the output size from $16 \times 16$ to $32 \times 32$. (iii) Instance segmentation performance saturates beyond an RoIAlign output size of $32 \times 32$. Taking into account the trade-off between performance and computational cost, we set RoIAlign output size as $32 \times 32$ by default.

\vspace{1mm}
\noindent\textbf{Training epochs.}
Table \ref{tab:effect_training_epochs} shows the effect of training epochs. We notice that doubling the number of training epochs only brings 0.3 gain in AP. This result suggests that our model achieves convergence promptly, which can effectively reduce the required training time.

\vspace{1mm}
\noindent\textbf{Large scale jittering.}
Table \ref{tab:effect_lsj} shows the effect of large scale jittering. We observe that using large scale jittering achieves a 0.3 AP improvement and increase the training GPU hours by 28.6\%. In light of the trade-off between training time and performance, we do not utilize large scale jittering in our following experiments.

\vspace{1mm}
\noindent\textbf{Instance mask head design.}
We compare the effect of mask head design in Table \ref{tab:effect_instance_mask_head}. Compared with segmenter head \cite{strudel2021segmenter} that contains linear projection and normalization, our simple dot product design achieves comparable AP scores with lower FLOPs and fewer number of parameters.

\begin{table}[t]
\begin{minipage}[t]{1\linewidth}
\vspace{2mm}
\centering
\setlength{\tabcolsep}{2pt}
\footnotesize
\renewcommand{\arraystretch}{1.3}
\resizebox{1.0\linewidth}{!}
{
\begin{tabular}{l|c|c|cccccc}
   mask head  & \#FLOPs$\blacktriangle$ & \#params$\blacktriangle$ & AP$^{\rm{mask}}$ & AP$^{\rm{mask}}_{50}$ & AP$^{\rm{mask}}_{75}$ & AP$^{\rm{mask}}_{\rm{S}}$ & AP$^{\rm{mask}}_{\rm{M}}$ & AP$^{\rm{mask}}_{\rm{L}}$  \\
    \shline
   \rowcolor{gray!10}dot product & $82.09$ G & $3.26$ M & $45.7$ & $67.5$ & $49.8$ & $25.6$ & $48.9$ & $64.1$ \\
   segmenter head  & $83.77$ G & $3.29$ M & $45.7$ & $67.6$ & $49.9$ & $25.6$ & $48.9$ & $64.1$ \\ 
\end{tabular}
}
\caption{\small{{
Effect of instance mask head design.}}
}
\label{tab:effect_instance_mask_head}
\end{minipage}
\end{table}

\begin{table}[t]
\begin{minipage}[t]{1\linewidth}
\centering
\setlength{\tabcolsep}{8pt}
\footnotesize
\renewcommand{\arraystretch}{1.3}
\resizebox{1.0\linewidth}{!}
{
\begin{tabular}{c|cccccc}
     batch size & AP$^{\rm{mask}}$ & AP$^{\rm{mask}}_{50}$ & AP$^{\rm{mask}}_{75}$ & AP$^{\rm{mask}}_{\rm{S}}$ & AP$^{\rm{mask}}_{\rm{M}}$ & AP$^{\rm{mask}}_{\rm{L}}$  \\
    \shline
   $2$ & $45.9$ & $67.7$ & $50.0$ & $25.7$ & $49.0$ & $64.2$ \\ 
   $4$ & $45.9$ & $67.6$ & $49.9$ & $25.8$ & $49.0$ & $64.3$ \\
   $8$ & $45.7$ & $67.5$ & $49.8$ & $25.6$ & $48.9$ & $64.1$ \\
\end{tabular}

}
\caption{\small{{
Effect of batch size.}}
}
\label{tab:effect_batch_size}
\end{minipage}
\end{table}

\begin{table}[t]
\begin{minipage}[t]{1\linewidth}
\centering
\setlength{\tabcolsep}{8pt}
\footnotesize
\renewcommand{\arraystretch}{1.2}
\resizebox{1.0\linewidth}{!}
{
\begin{tabular}{c|cccccc}
     layer\# & AP$^{\rm{mask}}$ & AP$^{\rm{mask}}_{50}$ & AP$^{\rm{mask}}_{75}$ & AP$^{\rm{mask}}_{\rm{S}}$ & AP$^{\rm{mask}}_{\rm{M}}$ & AP$^{\rm{mask}}_{\rm{L}}$  \\
    \shline
   $2$ & $45.7$ & $67.5$ & $49.8$ & $25.7$ & $48.9$ & $64.0$ \\ 
   $4$ & $45.7$ & $67.5$ & $49.8$ & $25.6$ & $48.9$ & $64.1$ \\
   $6$ & $45.5$ & $67.5$ & $49.3$ & $25.6$ & $48.7$ & $63.7$ \\
\end{tabular}

}
\caption{\small{{
Layer index of the encoder feature map $\mathbf{E}$.
}}
}
\label{tab:feat_fetch_layer}
\end{minipage}
\end{table}

\vspace{1mm}
\noindent\textbf{Batch size.}
Table \ref{tab:effect_batch_size} shows the effect of batch size. We observe that reducing the batch size from 8 to 4 or 2 even brings slightly improvements in performance (+0.2 AP). It is worth noting that our method can run on a single V100 GPU with 16G memory when the batch size is 2.
This highlights the effectiveness and computational resource efficiency of our approach.

\vspace{1mm}
\noindent\textbf{Layer index of the encoder feature map.}
We compare the encoder feature map $\mathbf{E}$ from different layers of the Transformer encoder in Table \ref{tab:feat_fetch_layer}. We notice that using the encoder feature map from shallower layers outperforms using that from the last layer of the encoder. We think this is because the feature map from the last layer of the encoder contains task-specific information relevant to detection, while feature maps from shallower layers contain more generalized object information that may facilitate segmentation. Therefore, we use the encoder feature map from layer\#$4$ by default.

\begin{table*}[h]
\begin{minipage}[t]{1\linewidth}
\vspace{2mm}
\centering
\setlength{\tabcolsep}{12pt}
\footnotesize
\renewcommand{\arraystretch}{1.35}
\resizebox{1.0\linewidth}{!}
{
\begin{tabular}{c|c|c|c|cccccc}
\# img. enc. layers & \# box enc. layers  & \#FLOPs$\blacktriangle$ & \#params$\blacktriangle$ & AP$^{\rm{mask}}$ & AP$^{\rm{mask}}_{50}$ & AP$^{\rm{mask}}_{75}$ & AP$^{\rm{mask}}_{\rm{S}}$ & AP$^{\rm{mask}}_{\rm{M}}$ & AP$^{\rm{mask}}_{\rm{L}}$ \\
\shline
$2$ & $2$  & $262.27$ G & $3.24$ M & $54.9$ & $78.9$ & $60.8$ & $37.2$ & $58.4$ & $72.9$ \\
$2$ & $3$  & $326.65$ G & $3.62$ M & $54.9$ & $78.9$ & $60.7$ & $37.0$ & $58.5$ & $72.9$  \\
\rowcolor{gray!10}$3$ & $2$  & $320.44$ G & $4.03$ M & $55.0$ & $78.9$ & $60.9$ & $37.1$ & $58.5$ & $73.0$  \\
$3$ & $3$ & $384.82$ G & $4.40$ M & $55.0$ & $78.9$ & $60.9$ & $37.2$ & $58.5$ & $73.1$ \\
$4$ & $4$ & $507.38$ G & $5.56$ M & $55.0$ & $78.9$ & $60.8$ & $37.2$ & $58.4$ & $73.2$ \\
\end{tabular}
}
\caption{\small{{
Effect of the depth of the image feature encoder and the box feature encoder on DINO + FocalNet-L.}}
}
\label{tab:feat_enc_depth_ablate}
\end{minipage}
\end{table*}

\begin{table}[t]
\begin{minipage}[t]{1\linewidth}
\vspace{-3mm}
\centering
\setlength{\tabcolsep}{0.25pt}
\footnotesize
\renewcommand{\arraystretch}{1.2}
\resizebox{1.0\linewidth}{!}
{
\begin{tabular}{l|ccc|cc|c|c}
detector method & \thead{image feat.\\ enc.} & \thead{box feat.\\ enc.} & \thead{query feat. \\ enc.} & partial finetune &  finetune & AP$^{\rm{mask}}$ & GPU Hours \\
\shline
\multirow{5}{*}{$\mathcal{H}$-DETR+R50} & \cmark & \cmark & \cmark & \xmark & \xmark & 45.7 & 49 \\ 
& \xmark & \xmark & \xmark & \cmark & \xmark &  43.8 & 92 \\
&\xmark & \xmark & \xmark & \xmark & \cmark & 43.9 & 99 \\
&\cmark & \cmark & \cmark & \cmark & \xmark & 45.6 & 100 \\
&\cmark & \cmark & \cmark & \xmark & \cmark & 46.0 & 108\\
\hline
\multirow{3}{*}{$\mathcal{H}$-DETR+Swin-L}  & \cmark & \cmark & \cmark & \xmark & \xmark & 54.0 & 172\\
& \cmark & \cmark & \cmark & \cmark & \xmark & 54.1 & 406\\
& \cmark & \cmark & \cmark & \xmark & \cmark & 54.1 & 532\\\hline
\multirow{3}{*}{DINO-DETR+FocalNet}  & \cmark & \cmark & \cmark & \xmark & \xmark & 54.9 & 136\\
& \cmark & \cmark & \cmark & \cmark & \xmark & 54.9 & 218 \\
& \cmark & \cmark & \cmark & \xmark & \cmark & 55.0 & 319\\
\end{tabular}
}
\caption{\footnotesize{
Effect of fine-tuning the whole DETR (fine-tune) or only the transformer encoder \& decoder within DETR (partial fine-tune). During fine-tuning, we set the learning rate of the original DETR parameters as $1/10$ of the ones of the additional new parameters.}}
\label{tab:finetune_detr}
\end{minipage}
\end{table}

\vspace{1mm}
\noindent\textbf{Effect of the depth of the image feature encoder and the box feature encoder}
Table \ref{tab:feat_enc_depth_ablate} shows the influence of the depth of image feature encoder and the box feature encoder on Mask Frozen-DINO-DETR with FocalNet-L as backbone. We observe that: (i) The instance segmentation performance of Frozen-DINO-DETR reaches saturation when the depth of the box feature encoder is 2, and further increasing its depth does not result in a performance gain. (ii) Increasing the depth of image feature encoder from 2 to 3 leads to a 0.1 increase in AP. Nevertheless, the performance saturates when the depth of image feature encoder is 3. Given these findings, we select the depth of the image feature encoder to be 3 and the depth of the box feature encoder to be 2 for the comparisons with state-of-the-art instance segmentation methods on the COCO test-dev.

\vspace{1mm}
\noindent\textbf{Effect of fine-tuning DETR}:
We further ablate the effect of fine-tuning the DETR-based object detector either entirely or partially, as outlined in Table~\ref{tab:finetune_detr}.
Accordingly,
we observe that (i) fine-tuning DETR brings consistent, albeit marginal, gains while significantly increasing the overall training GPU hours; (ii) our method achieves the best trade-off between performance and training cost.

\begin{figure}[t]
\centering
\includegraphics[width=0.5\textwidth]{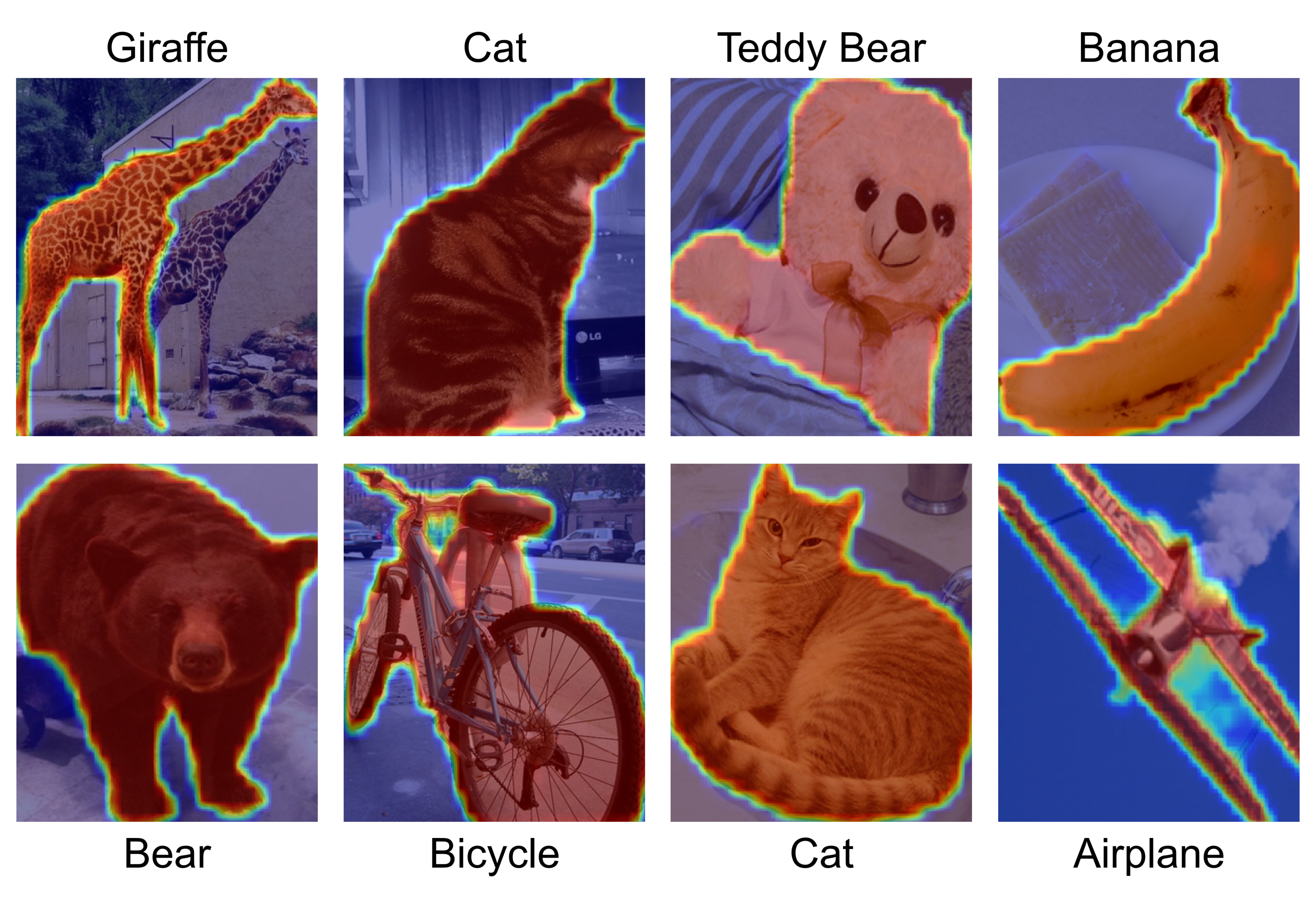}
\caption{\small{Visualizing the segmentation probability maps of our approach.}}
\label{fig:box2q_attn}
\vspace{-3mm}
\end{figure}

\vspace{1mm}
\noindent\textbf{Qualitative results.}
Figure~\ref{fig:box2q_attn} shows the instance segmentation probability maps based on our approach. We notice that the probability maps precisely capture the object boundaries, which support the strong performance of our approach on instance segmentation tasks.

\section{Conclusion}
In this work, we have presented the detailed techniques for converting an existing off-the-shelf DETR-based object detector into a strong instance segmentation model with minimal training time and resources. Our approach is remarkably simple yet effective. 
We verify the effectiveness of our approach by reporting state-of-the-art instance segmentation results while accelerating the training by more than $10\times$ times.
We believe our simple approach can inspire more research on advancing the state-of-the-art in instance segmentation model design.

{\small
\bibliographystyle{ieee}
\bibliography{egbib}
}

\end{document}